\documentclass[sigconf,nonacm]{acmart}

\usepackage{array,multirow,colortbl}
\usepackage{algorithm}
\usepackage[noend]{algpseudocode}
\usepackage[breakable,skins]{tcolorbox}
\definecolor{PromptBorder}{HTML}{B8C8D9}
\definecolor{PromptBackground}{HTML}{F5F7FA}
\definecolor{PromptHeading}{HTML}{36536F}
\newtcolorbox{promptbox}[1]{%
  enhanced,breakable,
  colback=PromptBackground,colframe=PromptBorder,
  colbacktitle=PromptBackground,coltitle=PromptHeading,
  fonttitle=\footnotesize\sffamily\mdseries,
  fontupper=\footnotesize\ttfamily,
  title={#1},title after break={#1\enspace (continued)},
  boxrule=0.4pt,arc=2pt,
  left=6pt,right=6pt,top=5pt,bottom=5pt,
  toptitle=5pt,bottomtitle=4pt,
  before skip=7pt,after skip=8pt,
  before upper={\raggedright\setlength{\parindent}{0pt}\setlength{\parskip}{3pt}}
}

\AtBeginDocument{}

\setcopyright{none}
\renewcommand\footnotetextcopyrightpermission[1]{}
\title{Pulse: Long-Horizon Tasks Need Streaming Personalization}

\author{Qijia Zhuang}
\affiliation{%
  \institution{Nanjing University}
  \country{}}

\author{Jialiang Liang}
\affiliation{%
  \institution{Xinjiang University}
  \country{}}

\author{Yuxiang Ren}
\affiliation{%
  \institution{Nanjing University}
  \country{}}
\renewcommand{\shortauthors}{Zhuang, Liang, and Ren}

\begin{document}
\setlength{\textfloatsep}{0pt plus 2pt}
\setlength{\dbltextfloatsep}{0pt plus 2pt}
\setlength{\floatsep}{0pt plus 2pt}
\setlength{\dblfloatsep}{0pt plus 2pt}
\setlength{\intextsep}{0pt plus 2pt}

\begin{abstract}
Large language model agents increasingly undertake complex, long-horizon tasks that benefit from sustained personalized support through long-term memory. Existing memory-augmented methods retrieve and use memory primarily from the initial request, with limited awareness of ongoing execution and emerging personalization needs. Agents may therefore overlook user preferences and project constraints during multistep tasks. We propose Pulse, a streaming personalization method for long-horizon tasks. An independent, lightweight memory agent continuously monitors execution, identifies emerging personalization needs and potential preference conflicts, and uses user and project memories to provide targeted interventions at appropriate stages. Reinforcement learning optimizes when and how the memory agent intervenes, improving the timeliness and accuracy of personalized support. Pulse scores 71.9\% on $\pi$-Bench and 46.4\% on MCP-Persona, achieving the highest overall scores among the compared methods on both datasets while improving task execution efficiency.
\end{abstract}


\maketitle

\section{Introduction}
\label{sec:introduction}

\begin{figure}[!t]
  \centering
  \includegraphics[width=\columnwidth]{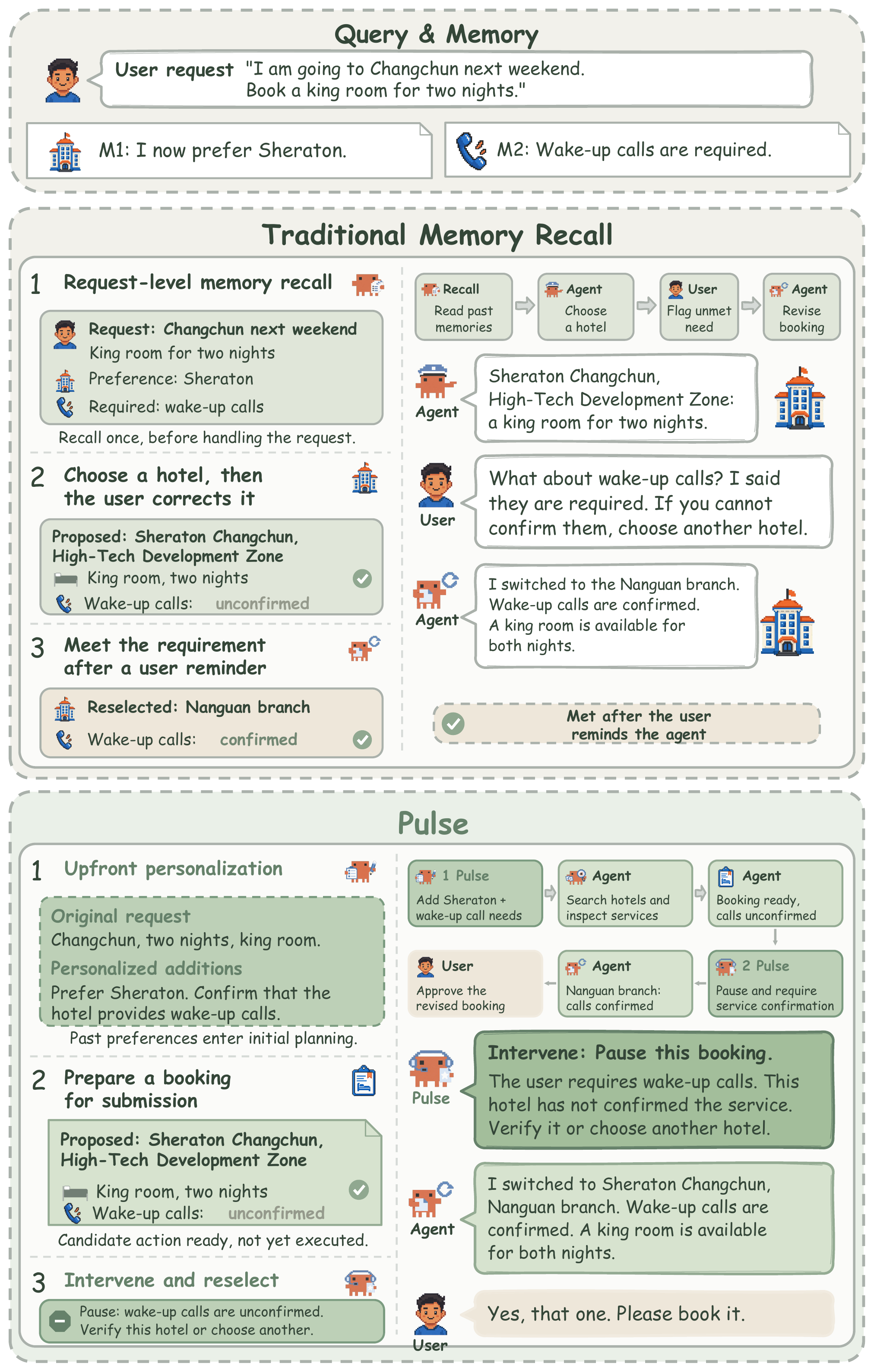}
  \caption{Hotel booking with request-level memory recall (top) and Pulse (bottom). The traditional workflow relies on a user reminder; Pulse prompts wake-up service verification or hotel reselection before submission.}
  \label{fig:intro-hotel-case}
  \Description{Vertically stacked hotel-booking workflows with the shared query and memories combined in one input panel above. The upper workflow selects a hotel with wake-up calls unconfirmed and revises it after a user reminder. The lower Pulse workflow prompts verification or reselection before submission.}
\end{figure}

As large language model agents take on long-horizon tasks such as online shopping, travel planning, and project collaboration, multi-step planning~\cite{yao2022react} and tool use~\cite{schick2023toolformer} become part of a continuous execution process. These tasks require agents to combine the current request with earlier interactions, user preferences, and evolving project state~\cite{park2023generativeagents}. Users rarely repeat every preference or established constraint~\cite{zhong2023memorybank}, so agents must apply long-term information at different stages to make decisions suited to the user~\cite{salemi2023lamp}.

User preferences may remain stable while their relevance to the current task becomes clear only during execution. The initial request typically describes a goal, whereas search, tool results, and intermediate decisions identify the concrete objects, available options, and conditions of execution. A memory that initially supplies background may therefore become a requirement to check once a specific action is proposed. Sustained personalization must identify when existing memory begins to affect the current decision and supply applicable preferences and constraints before the corresponding action is executed.

Existing memory methods extract, organize, and retrieve historical information to preserve background and preferences across sessions~\cite{wang2023longmem}. However, retrieval conditioned on the initial request struggles to cover new entities, subgoals, and candidate actions that emerge during execution~\cite{jiang2023flare}; even relevant memory included in the context may be overlooked in later decisions~\cite{liu2023lostmiddle}. This creates two challenges: \textbf{first, long-horizon execution requires sustained personalization}, with relevant memory supplied as the task state changes; \textbf{second, memory operations occupy the task agent's execution process}. Repeatedly asking the task agent to retrieve, filter, and check constraints adds reasoning turns and context load, extending the execution path of the main task.

We study sustained personalization for long-horizon tasks involving Web applications. Figure~\ref{fig:intro-hotel-case} illustrates online hotel booking: even after retrieving a preference for Sheraton and a requirement for wake-up calls, the agent may submit a booking without verifying the service. Pulse prompts service verification or hotel reselection before submission, allowing existing memory to influence the concrete decision in time.

To this end, we propose Pulse, a streaming personalization method supported by an independent memory agent. The memory agent augments the initial request using user and project memories, continuously monitors the task state during execution, and retrieves memory and selects evidence as needed. Based on the available evidence, it chooses silence, non-blocking supplementation, or pre-commit intervention, providing sustained memory support for the task agent's planning and tool execution.

Pulse must also decide whether retrieved evidence is sufficient to influence the current decision~\cite{asai2023selfrag}. Incorrect interruptions can trigger unnecessary replanning and disrupt otherwise valid execution. To strengthen the memory agent's judgment of memory applicability and intervention timing, we train its execution-time policy with Group Relative Policy Optimization (GRPO)~\cite{shao2024deepseekmath}. By rewarding correct memory-use actions and penalizing incorrect interventions, this training improves interruption precision and reduces the impact of incorrect interruptions on task execution.

Our contributions are threefold:
\begin{itemize}
\item We formulate sustained personalization for long-horizon tasks and design two complementary types of long-term memory: user Topic graphs encode preferences across tasks, while project memory preserves reusable workflows and value memories to support execution-time decisions.
\item We develop a streaming online personalization framework driven by an independent memory agent that gates retrieval with execution events and chooses silence, non-blocking supplementation, or pre-commit intervention. GRPO optimizes its memory-use policy.
\item Experiments on $\pi$-Bench and MCP-Persona show that Pulse improves personalized task performance in Web applications and effectively increases memory-use efficiency and task execution speed.
\end{itemize}

\section{Related Work}
\label{sec:related-work}

\subsection{Streaming Agents}
\label{sec:rw-streaming}

Streaming agents continuously receive new observations, update their context, and decide when to respond. Proactive assistance learns intervention opportunities from human activities and user feedback, linking response timing with users' willingness to accept assistance~\cite{lu2024proactive}. In continuous video interaction, compressed visual histories support real-time understanding~\cite{zhang2024flashvstream}, while temporally aligned training enables models to answer questions as new frames arrive~\cite{chen2024videollmonline}.

Streaming decisions also require coordination between response timing and evidence use. Decoupling perception, decision, and reaction supports asynchronous interaction~\cite{qian2025dispider}, while anticipating future evidence guides selective perception and historical retrieval~\cite{yang2025streamagent}. Bounded KV caches retain attention sinks and recent tokens to sustain continuous inference~\cite{xiao2023streamingllm}; corresponding evaluations examine both real-time perception and understanding of earlier events~\cite{lin2024streamingbench}. In Web interaction, page screenshots and interactive elements support action selection, and new observations after execution drive subsequent decisions~\cite{he2024webvoyager}. Feedback on state changes after actions helps verify execution outcomes~\cite{abuelsaad2024agente}, while hierarchical planning and local exploration support adjustments to subsequent steps based on new observations~\cite{zhang2024webpilot}. As the task state changes, systems must adjust their response timing and use of historical evidence.

Pulse brings streaming to personalized memory: the memory agent continuously observes Web task execution, supplies user and project evidence as needed, and intervenes before conflicting actions are submitted.

\subsection{Personalized Memory}
\label{sec:rw-memory}

Personalization infers preferences from interaction histories and adapts outputs accordingly. Prior work preserves information across sessions through long-term memory and selective forgetting~\cite{zhong2023memorybank}, and retrieves user profiles to support personalized generation~\cite{salemi2023lamp}. Dynamic profiles~\cite{li2024ldagent} and user-centered retrieval~\cite{zerhoudi2024personarag} further adapt to changing needs in long-term interaction, supplying relevant preferences for the current request.

Memory organization and updates affect how historical evidence is stored and used. External retrieval~\cite{lewis2020rag} and hierarchical management~\cite{packer2023memgpt} connect working context with long-term storage. Recursive summaries support retrieval at multiple levels~\cite{sarthi2024raptor}, while graph structures~\cite{gutierrez2024hipporag} and linked notes~\cite{xu2025amem} establish connections between pieces of information. Fact extraction~\cite{chhikara2025mem0}, topic summaries, and offline consolidation~\cite{fang2025lightmem} maintain memory content. Verbal feedback~\cite{shinn2023reflexion} and trajectory comparisons~\cite{zhao2023expel} help distill reusable experience, and experience stores further adjust reuse and prune outdated procedures according to utility~\cite{cao2025reme}. For Web tasks, prior work derives reusable workflows from task experience to guide subsequent execution~\cite{wang2024agentworkflowmemory}; reinforcement learning optimizes memory management and use policies~\cite{yan2025memoryr1}. Our Pulse combines personalized user and project memories, which the memory agent retrieves and uses as needed during execution, to provide sustained personalization for long-horizon tasks in Web applications.

\section{Method}
\label{sec:method}

\subsection{Problem Formulation and Overview}
\label{sec:method-overview}

We study long-horizon tasks whose personalization requirements become relevant during execution. A task agent plans, uses tools, and produces results. Pulse supplies applicable preferences and project constraints from historical interactions, deciding what evidence to deliver, when to deliver it, and whether supplementation or intervention is needed, while avoiding unnecessary retrieval and interruptions.

\begin{figure*}[t]
  \centering
  \includegraphics[width=\textwidth]{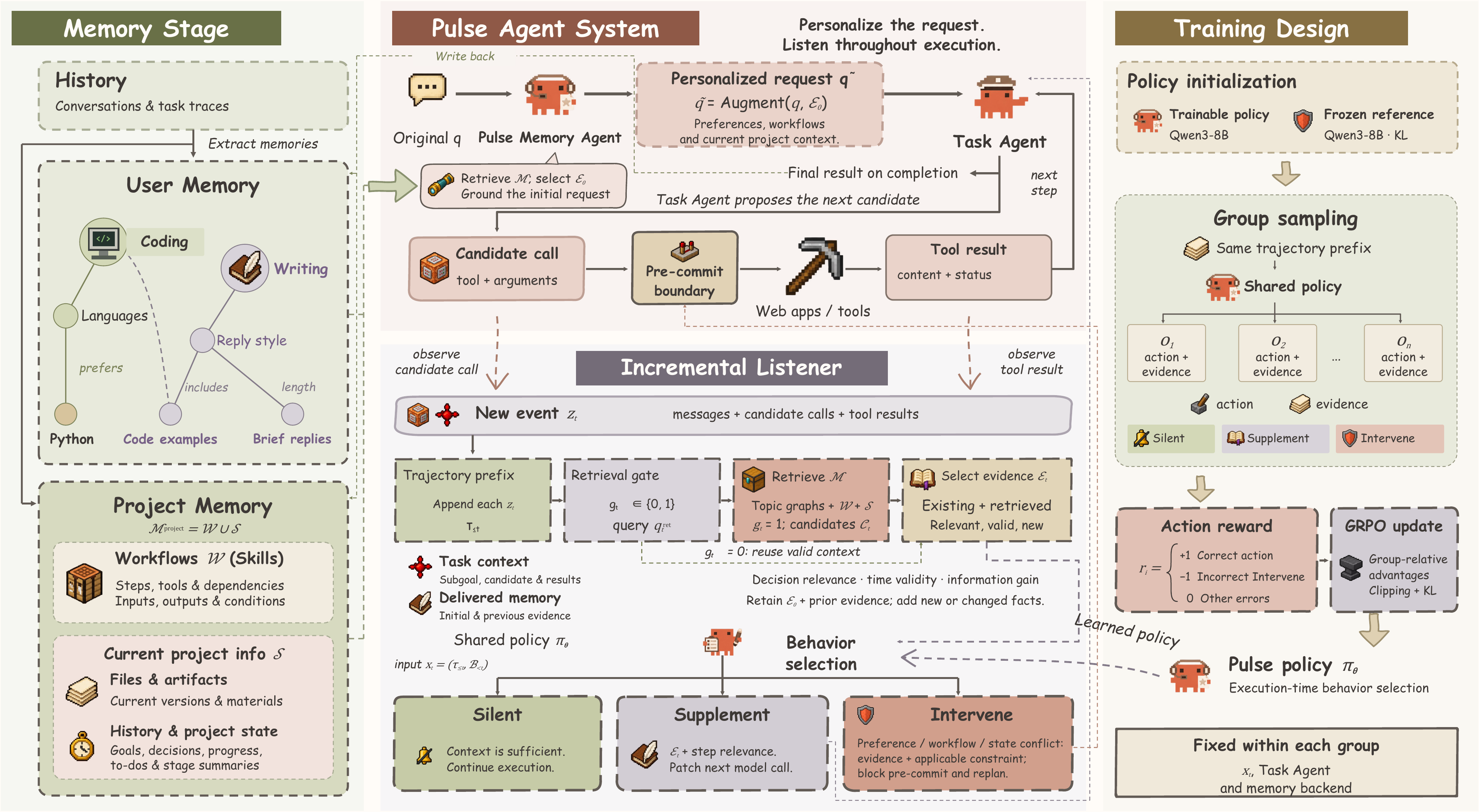}
  \caption{Overview of Pulse. Left: historical interactions and task traces form a user Topic graph and project memory containing workflows $\mathcal{W}$ and current information $\mathcal{S}$. Center: Pulse augments the request with initial evidence $\mathcal{E}_0$, incrementally observes messages, candidate calls, and tool results, gates retrieval, and selects silence, supplementation, or pre-commit intervention. Right: GRPO optimizes the execution-time policy using action correctness and an incorrect-intervention penalty.}
  \label{fig:pulse-framework}
  \Description{Three-part Pulse architecture. Historical conversations and task traces populate user Topic graphs and project workflows and state. One Pulse agent first personalizes the request, then listens to exposed execution events. A retrieval gate either reuses valid context or queries memory and selects evidence. Behavior selection yields Silent, Supplement for the next model call, or Intervene before a candidate tool call executes. GRPO trains the execution-time policy from a base model using group samples, a reward of plus one for a correct action, minus one for an incorrect intervention, and zero for other errors, with a frozen reference policy for KL regularization.}
\end{figure*}

An independent Pulse memory agent handles both initial personalization and execution-time monitoring. New entities, subgoals, and candidate actions can trigger retrieval; the policy selects silence, supplementation, or intervention. Figure~\ref{fig:pulse-framework} shows long-term memory, online personalization, and policy training.

Let $q$ denote the original request, $\widetilde q$ the request with initial memory, $\pi_{\mathrm{task}}$ the task policy, and $\pi_{\boldsymbol{\theta}}$ the shared Pulse policy with parameter vector $\boldsymbol{\theta}$. Events are indexed by $t=1,2,\ldots$, with initial prefix $\tau_{\leq0}=(q,\widetilde q)$. After event $t$, the observable prefix is
\begin{equation}
\tau_{\leq t}=(q,\widetilde q,z_1,\ldots,z_t),
\label{eq:trajectory}
\end{equation}
where $z_t$ contains a task-agent message, candidate tool call, or tool result; $\mathbin{\|}$ concatenates sequences, with $(z_t)$ a singleton sequence. Applicable memory at this prefix is
\begin{equation}
\mathcal{A}_t=\{m_j\in\mathcal{M}:\operatorname{Applies}(m_j,\tau_{\leq t})=1\}.
\label{eq:applicable-memory}
\end{equation}
Applicability requires the relevant user or project scope, temporal validity, and relevance to the current decision. A newly identified file or subgoal can change $\mathcal{A}_t$ even when the original request is unchanged. This set contains applicable records; the retrieval and selection process below determines which evidence to deliver.

The listener receives public events and appends tool names, arguments, and results to the prefix. Supplements enter the next task-agent decision. For a conflict, the runner pauses the candidate at the pre-commit boundary and returns evidence for replanning. Task completion triggers the memory updates in Section~\ref{sec:long-term-memory}.

\subsection{Long-Term Memory}
\label{sec:long-term-memory}

Long-term memory combines sets of retrievable user and project records, organized by the Topic graph and project structures:
\begin{equation}
\mathcal{M}=\mathcal{M}^{\mathrm{user}}\cup\mathcal{M}^{\mathrm{project}}.
\label{eq:memory-spaces}
\end{equation}

\paragraph{User memory.}
User memory takes the form of a Topic-managed multilayer knowledge graph. A thematic layer organizes domains into Topics and subtopics; an entity layer links people, objects, places, and activities; a factual layer stores attributes, preferences, and applicability conditions. Topics index these relations. In Figure~\ref{fig:pulse-framework}, Coding links Languages to Python, while Writing links Reply style to code examples and brief replies. Code examples connect both Topics for joint retrieval.

Retrieval is driven by the request or a new execution event. Pulse first locates Topics using the current subgoal, entities, and proposed action, then follows subtopic and entity relations to obtain facts together with their applicability conditions. Topic selection narrows the candidate space, and relational expansion recovers the context needed to interpret individual preferences. When explaining code, for example, the language preference can lead to requirements on examples and response length. Evidence selection considers step relevance, temporal validity, and information gain over the existing context, yielding facts and conditions that can affect the next decision.

Historical interactions initialize the graph, and new information from subsequent tasks is written into the relevant Topic subgraphs. Existing entities retain their identities; new facts attach to their entities and Topics, with cross-topic links for facts spanning multiple domains. An explicit preference revision records its source, update time, and relation to the superseded version, while task-specific requests retain their applicability scope. This organization accumulates long-term preferences while allowing retrieval to distinguish revised facts from currently valid constraints. Appendix~\ref{sec:appendix-memory-fill} and Appendix~\ref{sec:appendix-memory-update} detail memory population and updates.

\paragraph{Project memory.}
Project memory contains reusable workflow experience and current project information:
\begin{equation}
\mathcal{M}^{\mathrm{project}}=\mathcal{W}\cup\mathcal{S}.
\label{eq:project-memory}
\end{equation}
$\mathcal{W}$ stores Skill-like workflows: task steps, tool-use procedures, dependencies, inputs, outputs, and applicability conditions. These records provide operational experience for the current subgoal. $\mathcal{S}$ stores project objectives, accepted decisions, files and artifacts, progress, pending items, and stage summaries. Pulse matches the evolving execution state against both reusable workflows and current project facts.

Project identity and version information determine which records can support a decision. A reusable workflow may remain relevant across stages, but its inputs must be interpreted against the current files and accepted decisions in $\mathcal{S}$. Retrieval therefore supplies workflow steps together with applicable project state; temporary project choices are stored within their project scope.

Each retrievable item is represented as
\begin{equation}
m_j=(c_j,s_j,\nu_j,\eta_j),
\label{eq:memory-item}
\end{equation}
where $j$ indexes records, $c_j$ contains a graph fact, workflow fragment, or project state; $s_j\in\{\mathrm{user},\mathrm{project}\}$ identifies its category; scalar timestamp $\nu_j$ records its update time; and metadata record $\eta_j$ stores the Topic or project, provenance, and applicability conditions. Pulse selects evidence from these structures and composes a patch for the current step.

After a task, new user information updates its corresponding Topic subgraph. Reusable execution experience becomes workflow memory, while decisions, artifact changes, progress, and summaries update current project information. Writes first assign user and project scopes, merge duplicate facts, and establish revision links while retaining provenance and effective times. Tool receipts ground execution-state updates, and workflows retain step dependencies and applicability conditions for retrieval at similar subgoals.

\subsection{The Pulse Agent}
\label{sec:pulse-agent}

\subsubsection{Initial Request Personalization}
\label{sec:initial-personalization}

At initialization, Pulse selects evidence $\mathcal{E}_0$ from the user Topic graph and project memory according to the original request $q$, forming
\begin{equation}
\widetilde q=\operatorname{Augment}(q,\mathcal{E}_0).
\label{eq:personalized-request}
\end{equation}
This evidence supplies preferences, workflows, and project context for initial planning and provides the context and deduplication reference for subsequent monitoring.

Initial evidence covers requirements identifiable from the request. Later tool results may introduce a concrete recipient, file, or artifact whose associated constraints become relevant only then. Execution-time retrieval connects those newly exposed choices to long-term memory.

\subsubsection{Incremental Observation and Retrieval}
\label{sec:incremental-retrieval}

During execution, Pulse receives event $z_t$. Let $\mathcal{B}_{<t}$ contain evidence actually delivered before event $t$, including $\mathcal{E}_0$. The controller input is $x_t=(\tau_{\leq t},\mathcal{B}_{<t})$. The memory agent decides whether to retrieve and generates a query:
\begin{equation}
\begin{aligned}
(g_t,q_t^{\mathrm{ret}})&=\pi_{\boldsymbol{\theta}}^{\mathrm{access}}(x_t),\\
g_t&\in\{0,1\}.
\end{aligned}
\label{eq:retrieval-gate}
\end{equation}
$\mathrm{access}$ and $\mathrm{action}$ denote the retrieval and action-selection calls to the same memory agent. New subgoals, entities, proposals, or state changes update memory needs. When $g_t=1$, Pulse performs
\begin{equation}
\mathcal{C}_t=\operatorname{Search}(\mathcal{M},q_t^{\mathrm{ret}}).
\label{eq:memory-search}
\end{equation}
Queries locate user Topic subgraphs, workflows, or current project information, yielding $\mathcal{C}_t\subseteq\mathcal{M}$; set $\mathcal{C}_t=\varnothing$ when $g_t=0$. Selected evidence satisfies $\mathcal{E}_t\subseteq(\mathcal{B}_{<t}\cup\mathcal{C}_t)\cap\mathcal{A}_t$. Pulse selects evidence according to current relevance, temporal validity, and information gain, reusing valid previous evidence to check candidate actions. Evidence delivered at this event enters the delivered-evidence snapshots of subsequent events.

Previously delivered facts can be reused when they remain applicable to the current step. Pulse selects patches according to evidence validity, project scope, and information gain. Retrieval and behavior selection are independent: existing evidence can directly support intervention, while newly retrieved results may lead the policy to remain silent.

For API-based task agents, the runner exposes candidate calls before tool execution and results afterward. Candidate arguments identify relevant preferences and project constraints, while tool results introduce new entities, artifacts, and execution states for the next retrieval decision.

\subsubsection{Supplementation and Pre-commit Intervention}
\label{sec:supplement-intervene}

Given $x_t$ and $\mathcal{E}_t$, Pulse selects $u_t$ over the event-dependent admissible action set $\mathcal{U}_t$:
\begin{equation}
\mathcal{U}_t=
\begin{cases}
\{\bot,\textsc{Supplement},\textsc{Intervene}\}, & \text{uncommitted call},\\
\{\bot,\textsc{Supplement}\}, & \text{other events}.
\end{cases}
\label{eq:pulse-actions}
\end{equation}
Here $u_t=\pi_{\boldsymbol{\theta}}^{\mathrm{action}}(x_t,\mathcal{E}_t;\mathcal{U}_t)$, choosing an action from $\mathcal{U}_t$; $\bot$ is the \textsc{Silent} action in Figure~\ref{fig:pulse-framework}. Pulse remains silent when the available context supports the decision. When memory supplies information needed for the next step, it selects \textsc{Supplement} and forms
\begin{equation}
p_t=\bigl(\mathcal{E}_t,\operatorname{Reason}(\mathcal{E}_t,x_t)\bigr).
\label{eq:memory-patch}
\end{equation}
The patch contains evidence and its relation to the current step and is added to the task agent's context at the next model call. When a proposal conflicts with user preferences, workflow requirements, or current project state, Pulse selects \textsc{Intervene} and returns the conflicting evidence and the constraint to revise. The runner pauses that candidate call, and the task agent replans. Superseded files, completed operations, and updated project decisions are examples of memory that can change an action choice. This corresponds to the pre-commit boundary in Figure~\ref{fig:pulse-framework}: the conflict is checked before the current call changes tool or environment state. Returned tool results become new events for subsequent monitoring and retrieval.

Intervention evidence must apply to the current candidate action and identify the argument, step dependency, or accepted decision that requires revision. Evidence that clarifies subsequent choices is delivered as a supplement.

Algorithm~\ref{alg:pulse-execution} summarizes Pulse's execution loop. The task agent generates events while Pulse independently selects evidence and a response. \textsc{Intervene} is available only at an uncommitted candidate call; messages and tool results permit silence or supplementation. Memory updates follow task completion.

\begin{algorithm}[t]
\caption{Pulse streaming personalization}
\label{alg:pulse-execution}
\small
\begin{algorithmic}
\Require Request $q$, memory $\mathcal{M}$, policies $\pi_{\mathrm{task}},\pi_{\boldsymbol{\theta}}$
\Ensure Task result $y$ and updated memory $\mathcal{M}$
\State $\mathcal{E}_0\gets\operatorname{Select}(\operatorname{Search}(\mathcal{M},q),q)$
\State $\widetilde q\gets\operatorname{Augment}(q,\mathcal{E}_0)$; $t\gets0$; $\tau_{\leq0}\gets(q,\widetilde q)$
\State $\mathcal{B}_{<1}\gets\mathcal{E}_0$; start the task agent and Pulse listener
\While{the task is unfinished}
    \State $t\gets t+1$; observe $z_t$; $\tau_{\leq t}\gets\tau_{\leq t-1}\mathbin{\|}(z_t)$
    \State Refresh $\mathcal{B}_{<t}$ from deliveries completed before $z_t$
    \State $x_t\gets(\tau_{\leq t},\mathcal{B}_{<t})$
    \State Predict $(g_t,q_t^{\mathrm{ret}})$; $\mathcal{C}_t\gets\varnothing$
    \If{$g_t=1$}
        \State $\mathcal{C}_t\gets\operatorname{Search}(\mathcal{M},q_t^{\mathrm{ret}})$
    \EndIf
    \State $\mathcal{E}_t\gets\operatorname{SelectValid}(\mathcal{B}_{<t},\mathcal{C}_t;x_t)$
    \State Select $u_t$ from $\mathcal{U}_t$
    \If{$u_t=\textsc{Supplement}$}
        \State Queue $p_t$ for the next task-agent call
    \ElsIf{$u_t=\textsc{Intervene}$}
        \State Block the candidate; return evidence for replanning
    \EndIf
\EndWhile
\State $y\gets$ final task result
\State $\mathcal{M}\gets\operatorname{WriteBack}(\mathcal{M},\tau_{\leq t},y)$
\State \Return $(y,\mathcal{M})$
\end{algorithmic}
\end{algorithm}

\subsection{Policy Optimization}
\label{sec:policy-optimization}

Pulse is initialized from Qwen3-8B and optimized for 100 updates using GRPO~\cite{shao2024deepseekmath}. Training compares alternative memory responses at the same decision point: the controller state $x_t$, valid memory-backend snapshot, and task agent remain fixed within each group. We sample
\begin{equation}
o_i\sim\pi_{\mathrm{old}}(\cdot\mid x_t),\qquad i=1,\ldots,n,\quad n\geq2.
\label{eq:group-sampling}
\end{equation}
Each $o_i$ contains an action and supporting evidence under the fixed access/retrieval protocol. The group-generating behavior policy $\pi_{\mathrm{old}}$ is frozen for the objective.

Labels depend on the current prefix and valid evidence. Silence is correct when no memory response is needed; \textsc{Supplement} is correct when evidence supports the next decision without blocking; \textsc{Intervene} is correct when an applicable conflict requires reconsidering the candidate call. The reward is
\begin{equation}
r_i=
\begin{cases}
1, & \text{if the Pulse action in } o_i \text{ is correct},\\
-1, & \text{if } o_i \text{ incorrectly selects Intervene},\\
0, & \text{otherwise}.
\end{cases}
\label{eq:reward}
\end{equation}
The $-1$ penalty targets interventions at states where silence or supplementation is appropriate. Such interruptions block otherwise valid progress and incur replanning or extra model calls. Other incorrect choices receive zero reward. This asymmetry trains the policy to distinguish useful memory support from unnecessary interruption.

We optimize the candidate-level GRPO surrogate
\begin{equation}
\begin{aligned}
\ell_{\mathrm{GRPO}}(\boldsymbol{\theta})
&=-\frac{1}{n}\sum_{i=1}^{n}\ell_i(\boldsymbol{\theta})\\
&\quad+\beta D_{\mathrm{KL}}\bigl(
\pi_{\boldsymbol{\theta}}(\cdot\mid x_t)
\Vert\pi_{\mathrm{ref}}(\cdot\mid x_t)\bigr),\\
\ell_i(\boldsymbol{\theta})
&=\min\bigl\{\rho_i(\boldsymbol{\theta})a_i,\\
&\quad\operatorname{clip}\bigl(\rho_i(\boldsymbol{\theta}),
1-\epsilon_{\mathrm{clip}},1+\epsilon_{\mathrm{clip}}\bigr)
a_i\bigr\}.
\end{aligned}
\label{eq:grpo}
\end{equation}
Here $a_i$ is the group-relative reward advantage and $\rho_i$ compares current and behavior-policy likelihoods. The clipped term favors higher-reward responses to the same controller state, making unnecessary \textsc{Intervene} less competitive than appropriate silence or supplementation under Equation~\ref{eq:reward}. KL regularization, weighted by $\beta$, anchors the policy to the independently frozen initial reference $\pi_{\mathrm{ref}}$. Advantage normalization, likelihood ratios, the clipping threshold, and training details are given in Appendix~\ref{sec:appendix-training-details}.

\section{Experiments}
\label{sec:experiments}

\subsection{Experimental Details}
\label{sec:exp-setup}

\begin{table*}[t]
\centering
\caption{$\pi$-Bench results. COMP is reported as a percentage; Store and Retrieve are thousands of tokens per task. \textcolor[HTML]{B42318}{Red} and \textcolor[HTML]{175CD3}{blue} mark the best and second-best values, respectively. Time is seconds per task.}
\label{tab:main-results}
\label{tab:main-pi}
\scriptsize
\setlength{\tabcolsep}{4pt}
\resizebox{\textwidth}{!}{%
\begin{tabular}{llrrrrrrrrr}
\toprule
Group & Method & \multicolumn{5}{c}{User Domain} & Overall & \multicolumn{2}{c}{Token Usage} &  \\
\cmidrule(lr){3-7}\cmidrule(lr){9-10}
 & & Researcher & Marketer & Pharmacist & Law Trainee & Financier & COMP  & Store Token (k) & Retrieve Token (k) & Time (s) \\
\midrule
Memory & Mem0 & 51.3 & 53.0 & 53.6 & 46.6 & 50.9 & 50.5 & 5.6 & 3.7 & 242.1 \\
 & LightMem & 50.6 & 40.9 & 61.8 & 47.7 & 53.8 & 51.6 & \textcolor[HTML]{B42318}{4.8} & \textcolor[HTML]{B42318}{3.1} & 277.3 \\
 & A-MEM & 45.5 & 48.5 & 48.2 & 48.3 & 53.8 & 49.0 & 8.3 & 6.1 & 243.2 \\
\midrule
Harness & OpenClaw & 56.4 & 42.4 & \textcolor[HTML]{B42318}{88.2} & 60.9 & 63.9 & 63.3 & 5.5 & 5.0 & 275.9 \\
 & dsh & 46.8 & 47.0 & 67.3 & \textcolor[HTML]{175CD3}{70.1} & 56.8 & 58.7 & 6.4 & 5.9 & 264.0 \\
 & ZCode & 57.7 & 50.0 & 76.4 & 67.2 & 58.0 & 62.5 & 5.9 & 5.7 & 288.5 \\
\midrule
Ours & Pulse w/o RL & \textcolor[HTML]{175CD3}{61.5} & \textcolor[HTML]{175CD3}{56.1} & 70.0 & \textcolor[HTML]{175CD3}{70.1} & \textcolor[HTML]{175CD3}{69.2} & \textcolor[HTML]{175CD3}{66.5} & \textcolor[HTML]{175CD3}{5.3} & \textcolor[HTML]{175CD3}{3.6} & \textcolor[HTML]{175CD3}{233.9} \\
 & Pulse (Full) & \textcolor[HTML]{B42318}{68.6} & \textcolor[HTML]{B42318}{59.1} & \textcolor[HTML]{175CD3}{80.9} & \textcolor[HTML]{B42318}{74.1} & \textcolor[HTML]{B42318}{71.6} & \textcolor[HTML]{B42318}{71.9} & \textcolor[HTML]{175CD3}{5.3} & 3.9 & \textcolor[HTML]{B42318}{203.8} \\
\bottomrule
\end{tabular}
}
\end{table*}

\begin{table*}[t]
\centering
\caption{MCP-Persona results. Acc is reported as a percentage; Store and Retrieve are thousands of tokens per task. \textcolor[HTML]{B42318}{Red} and \textcolor[HTML]{175CD3}{blue} mark the best and second-best values, respectively. Time is seconds per task.}
\label{tab:main-mcp}
\scriptsize
\setlength{\tabcolsep}{4pt}
\resizebox{\textwidth}{!}{%
\begin{tabular}{llrrrrrrrrrrr}
\toprule
Group & Method & \multicolumn{4}{c}{Single-Server} & \multicolumn{3}{c}{Cross-Server} & Overall & \multicolumn{2}{c}{Token Usage} &  \\
\cmidrule(lr){3-6}\cmidrule(lr){7-9}\cmidrule(lr){11-12}
 & & CP & CM & SM & EM & Lark & Rednote & Hodgepodge & Acc  & Store Token (k) & Retrieve Token (k) & Time (s) \\
\midrule
Memory & Mem0 & 32.6 & 18.0 & 33.1 & 43.1 & 28.6 & 25.5 & 6.3 & 29.5 & 8.0 & 3.4 & 181.1 \\
 & LightMem & 35.2 & 20.7 & 35.5 & 45.1 & 31.6 & 27.7 & 18.8 & 32.2 & \textcolor[HTML]{B42318}{3.8} & \textcolor[HTML]{B42318}{1.9} & 139.8 \\
 & A-MEM & 27.9 & 14.7 & 28.5 & 39.2 & 23.6 & 21.2 & 6.3 & 25.0 & 5.3 & 4.1 & 164.6 \\
\midrule
Harness & OpenClaw & 37.8 & 19.3 & 37.8 & 47.1 & 34.3 & 30.4 & 18.8 & 34.3 & 4.9 & 3.1 & 157.8 \\
 & dsh & 40.4 & 21.3 & 40.1 & 46.1 & 37.0 & 32.6 & 18.8 & 36.5 & 5.3 & 3.7 & 147.3 \\
 & ZCode & 39.8 & 23.3 & 39.5 & 49.0 & 35.2 & 28.8 & 25.0 & 35.8 & 6.2 & 4.3 & 166.9 \\
\midrule
Ours & Pulse w/o RL & \textcolor[HTML]{175CD3}{43.8} & \textcolor[HTML]{175CD3}{24.7} & \textcolor[HTML]{175CD3}{44.2} & \textcolor[HTML]{175CD3}{53.9} & \textcolor[HTML]{175CD3}{41.1} & \textcolor[HTML]{175CD3}{39.1} & \textcolor[HTML]{175CD3}{37.5} & \textcolor[HTML]{175CD3}{41.1} & \textcolor[HTML]{175CD3}{4.7} & 3.3 & \textcolor[HTML]{175CD3}{128.7} \\
 & Pulse (Full) & \textcolor[HTML]{B42318}{48.7} & \textcolor[HTML]{B42318}{30.7} & \textcolor[HTML]{B42318}{49.4} & \textcolor[HTML]{B42318}{61.8} & \textcolor[HTML]{B42318}{47.0} & \textcolor[HTML]{B42318}{41.8} & \textcolor[HTML]{B42318}{43.8} & \textcolor[HTML]{B42318}{46.4} & \textcolor[HTML]{175CD3}{4.7} & \textcolor[HTML]{175CD3}{2.6} & \textcolor[HTML]{B42318}{118.3} \\
\bottomrule
\end{tabular}
}
\end{table*}

Evaluation configurations, per-category ablation results, and supplementary experiments appear in Appendix~\ref{sec:appendix-experiments}.

\subsubsection{Datasets}

$\pi$-Bench~\cite{zhang2026pibench} combines Web application tools with the work requirements of five user roles---researchers, marketers, pharmacists, law trainees, and finance professionals---to construct long-horizon workflows with cross-session dependencies. Tasks unfold in persistent project workspaces and require reusing historical preferences and project state in subsequent operations. We report COMP for task completeness.

MCP-Persona~\cite{wang2026mcppersona} covers personal applications for collaboration, content management, social media, and email, using environment simulation to construct single- and cross-service tasks. Agents must combine user profiles, preferences, and application state to complete multistep operations. We report Acc for checkpoint completion.

Persona2Web~\cite{kim2026persona2web} targets open Web environments, providing user histories that implicitly reveal preferences and ambiguous requests that require agents to infer user intent from history. We use its data-generation scripts to generate an entirely new training set for GRPO training of Pulse's execution-time policy, and use the benchmark's existing data for supplementary experiments. Supplementary evaluation appears in Appendix~\ref{sec:appendix-persona2web}.

\subsubsection{Comparison Methods}

As agent harnesses integrate persistent memory into session management, context organization, and tool invocation, the execution framework also affects how memory is stored and used. We therefore compare two groups---traditional memory systems and harness memory strategies---to examine independent memory components and memory mechanisms embedded in execution workflows.

\paragraph{Traditional memory systems.}
Mem0~\cite{chhikara2025mem0} maintains long-term memory through fact extraction, updates, and consolidation. LightMem~\cite{fang2025lightmem} combines lightweight filtering, topic aggregation, and offline updates to reduce memory-processing costs. A-MEM~\cite{xu2025amem} organizes historical information through structured notes and evolving links.

\paragraph{Harness memory strategies.}
\href{https://docs.openclaw.ai/concepts/memory}{OpenClaw} integrates persistent memory, workspace files, and skills into the execution environment. \href{https://github.com/deepseek-ai/deepseek-harness}{dsh} composes the agent loop, tools, and session management through plugins. \href{https://zcode.z.ai/en/docs/agent-framework}{ZCode} maintains project files, execution history, and tool context across long-running tasks.

Pulse w/o RL retains the full memory architecture and execution-time interventions without GRPO updates. Pulse (Full) uses GRPO to optimize the complete control policy.

\subsubsection{Implementation Settings}

We use GPT-5.4 mini as the task agent and text-embedding-3-small for embeddings. All memory agents across the compared methods use Qwen3-8B as their backbone. Pulse is initialized from the base policy and trained with GRPO for 100 steps with a batch size of 32. Store Token counts the memory model's token output to the memory system; Retrieve Token counts the memory tokens it supplies to the task model.

Time is the mean end-to-end duration per task in seconds, covering the task model, memory operations, and tool calls.

\subsection{Main Results}
\label{sec:exp-main}

\paragraph{$\pi$-Bench.}
\looseness=-1
Pulse (Full) achieves \textbf{71.9\% overall COMP}, exceeding OpenClaw, the strongest harness baseline, by 8.6 percentage points and LightMem, the strongest memory baseline, by 20.3 points. The category results place Pulse first in research, marketing, law, and finance, with scores of 68.6\%, 59.1\%, 74.1\%, and 71.6\%, respectively. OpenClaw retains the highest pharmacy score at 88.2\%, compared with 80.9\% for Pulse, showing variation across domains alongside the overall improvement. Compared with Pulse w/o RL, the full model increases overall COMP from 66.5\% to 71.9\%, a gain of 5.3 points, with improvements in all five domains. Research improves by 7.1 points, pharmacy by 10.9 points, and marketing by 3.0 points.

\begin{table*}[t]
\centering
\caption{Component ablations. Store and Retrieve are k tokens per task; Time is seconds per task.}
\label{tab:ablations}
\footnotesize
\definecolor{AblationImprovement}{HTML}{308653}
\definecolor{AblationRegression}{HTML}{C44F4F}
\newcommand{\ablationdelta}[3]{\mbox{#1\textsuperscript{\scriptsize\color{#3}$#2$}}}
\setlength{\tabcolsep}{2.5pt}
\renewcommand{\arraystretch}{1.16}
\begin{tabular}{p{0.072\textwidth}p{0.205\textwidth}*{8}{>{\centering\arraybackslash}p{\dimexpr(\textwidth-0.277\textwidth-20\tabcolsep)/8\relax}}}
\toprule[0.8pt]
\multirow{2}{*}{Group} & \multirow{2}{*}{Variant} & \multicolumn{4}{c}{$\pi$-Bench} & \multicolumn{4}{c}{MCP-Persona} \\
\cmidrule(lr){3-6}\cmidrule(lr){7-10}
 & & COMP (\%) & Store (k) & Retrieve (k) & Time (s) & Acc (\%) & Store (k) & Retrieve (k) & Time (s) \\
\midrule[0.45pt]
\multirow{4}{*}{Memory} & Flat Memory & 49.9 & 7.4 & 4.3 & 228.4 & 27.9 & 7.3 & 3.1 & 143.2 \\
 & Flat + User & \ablationdelta{53.6}{+3.7}{AblationImprovement} & \ablationdelta{6.7}{-0.7}{AblationImprovement} & \ablationdelta{4.2}{-0.1}{AblationImprovement} & \ablationdelta{248.5}{+20.1}{AblationRegression} & \ablationdelta{30.2}{+2.3}{AblationImprovement} & \ablationdelta{6.8}{-0.6}{AblationImprovement} & \ablationdelta{2.9}{-0.2}{AblationImprovement} & \ablationdelta{158.6}{+15.4}{AblationRegression} \\
 & Flat + Project & \ablationdelta{54.4}{+4.4}{AblationImprovement} & \ablationdelta{6.5}{-0.9}{AblationImprovement} & \ablationdelta{4.1}{-0.2}{AblationImprovement} & \ablationdelta{245.7}{+17.3}{AblationRegression} & \ablationdelta{30.2}{+2.3}{AblationImprovement} & \ablationdelta{6.6}{-0.7}{AblationImprovement} & \ablationdelta{2.8}{-0.3}{AblationImprovement} & \ablationdelta{154.3}{+11.1}{AblationRegression} \\
 & Flat Memory + Pulse w/o RL & \ablationdelta{61.2}{+11.3}{AblationImprovement} & 7.4 & \ablationdelta{4.6}{+0.3}{AblationRegression} & \ablationdelta{202.7}{-25.7}{AblationImprovement} & \ablationdelta{35.9}{+8.0}{AblationImprovement} & 7.3 & \ablationdelta{3.2}{+0.1}{AblationRegression} & \ablationdelta{124.4}{-18.8}{AblationImprovement} \\
\arrayrulecolor[HTML]{D4DCE5}\cmidrule(l){2-10}\arrayrulecolor{black}
\multirow{2}{*}{Retrieval} & Start-only & 56.4 & 5.3 & 3.5 & 215.7 & 33.1 & 5.3 & 2.2 & 110.4 \\
 & Streaming supplement-only & \ablationdelta{61.8}{+5.3}{AblationImprovement} & 5.3 & \ablationdelta{4.2}{+0.7}{AblationRegression} & \ablationdelta{229.6}{+13.9}{AblationRegression} & \ablationdelta{36.3}{+3.2}{AblationImprovement} & 5.3 & \ablationdelta{2.9}{+0.6}{AblationRegression} & \ablationdelta{137.6}{+27.2}{AblationRegression} \\
\arrayrulecolor[HTML]{D4DCE5}\cmidrule(l){2-10}\arrayrulecolor{black}
Training & w/o GRPO & 66.5 & 5.3 & 3.6 & 233.9 & 41.1 & 4.7 & 3.3 & 128.7 \\
\addlinespace[3pt]
\multicolumn{2}{l}{\textbf{Full Pulse}} & \ablationdelta{71.9}{+5.3}{AblationImprovement} & 5.3 & \ablationdelta{3.9}{+0.3}{AblationRegression} & \ablationdelta{203.8}{-30.1}{AblationImprovement} & \ablationdelta{46.4}{+5.3}{AblationImprovement} & 4.7 & \ablationdelta{2.6}{-0.7}{AblationImprovement} & \ablationdelta{118.3}{-10.4}{AblationImprovement} \\
\bottomrule[0.8pt]
\end{tabular}
\par\vspace{4pt}
\begin{minipage}{\textwidth}
\footnotesize\raggedright
\textit{Note.} Memory variants are compared with Flat Memory; Retrieval with Start-only; Full Pulse with w/o GRPO. Units are percentage points, k tokens, and seconds. \textcolor{AblationImprovement}{Green} indicates improvement and \textcolor{AblationRegression}{red} indicates regression; higher performance and lower Token and Time values are better.
\end{minipage}
\end{table*}

\paragraph{MCP-Persona.}
\looseness=-1
Pulse (Full) achieves \textbf{46.4\% overall Acc}, exceeding dsh, the strongest harness baseline, by 9.9 percentage points and LightMem, the strongest memory baseline, by 14.2 points. Pulse leads all seven categories, covering both single-service and cross-service tasks. EM reaches 61.8\%, while the cross-service Lark and Rednote categories reach 47.0\% and 41.8\%, compared with 37.0\% and 32.6\% for dsh. Compared with Pulse w/o RL, overall Acc increases from 41.1\% to 46.4\%, a gain of 5.3 points, with improvements in every category. Among cross-service categories, Lark improves by 5.9 points and Rednote by 2.7 points. Within the single-service group, EM improves from 53.9\% without RL to 61.8\% with the full model, a gain of 7.8 points, complementing the cross-service gains.

\paragraph{Memory-use efficiency.}
\looseness=-1
Higher task completion accompanies fewer memory tokens supplied to the task model relative to the strongest harnesses. On $\pi$-Bench, Pulse reduces Retrieve Token from OpenClaw's 5.0k to 3.9k, a reduction of 22.0\%; on MCP-Persona, it reduces dsh's 3.692k to 2.605k, a reduction of \textbf{29.4\%}. LightMem has lower Store and Retrieve costs on both benchmarks, representing a different tradeoff between task quality and memory cost. Comparing the two Pulse policy configurations, Store Token is 4.7k for both on MCP-Persona; Retrieve Token changes differ: $\pi$-Bench increases from 3.6k to 3.9k, whereas MCP-Persona decreases from 3.287k to 2.605k, a reduction of 20.7\%.

\paragraph{Runtime.}
\looseness=-1
Pulse runs faster than the other baselines because its memory agent automatically supplements long-horizon execution, avoiding serial retrieval waits for the task agent. GRPO further reduces runtime from 233.9 to 203.8 seconds on $\pi$-Bench and from 128.7 to 118.3 seconds on MCP-Persona. More accurate memory retrieval and fewer unnecessary interruptions jointly improve task performance and execution time.

\subsection{Ablation Studies}
\label{sec:exp-ablation}

We organize eight ablation settings into three groups covering memory organization, execution-time support, and RL optimization, with Full Pulse as the shared reference.

\paragraph{Memory organization.}
Four settings retain the same user and project information and vary its organization and retrieval:
\begin{itemize}
\item \emph{Flat Memory}: both user and project memories use flat text, retaining facts, provenance, timestamps, and applicability conditions.
\item \emph{Flat + User}: user memory uses the Topic graph, while project memory remains flat.
\item \emph{Flat + Project}: user memory remains flat, while project memory is organized into workflows and current project information.
\item \emph{Full Pulse}: both the user Topic graph and the project organization into workflows and current information are enabled.
\end{itemize}

\paragraph{Execution-time support.}
We fix the full memory structure and initial evidence. \emph{Start-only} supplies memory at initialization and disables subsequent retrieval, supplementation, and intervention. \emph{Streaming supplement-only} retains execution-time retrieval and selective delivery, converting blocking decisions into advisory supplements. Full Pulse additionally enables pre-commit intervention.

\paragraph{RL optimization.}
\emph{w/o GRPO} retains the full architecture and uses the initial backbone policy without RL updates, allowing comparison with Full Pulse to assess policy optimization. \emph{Flat Memory + Pulse w/o RL} uses flat user and project memories while retaining Pulse's execution-time retrieval, selective supplementation, and pre-commit intervention under the initial policy without GRPO updates. Comparison with Flat Memory assesses the effect of adding Pulse's execution-time support to flat memory; comparison with w/o GRPO assesses memory organization under the untrained policy.

Per-category completed counts, percentages, and costs for each setting appear in Appendix Tables~\ref{tab:ablation-details-pi} and~\ref{tab:ablation-details-mcp}.

\looseness=-1
Both single-structure variants reduce Store Token and Retrieve Token and outperform Flat Memory on both benchmarks. Full Pulse achieves the strongest performance when both structures are enabled.

\looseness=-1
Streaming supplement-only outperforms Start-only on both benchmarks, showing the benefit of updating evidence during execution. Full Pulse gains another 10.1 percentage points on each benchmark while delivering fewer memory tokens. This supports using pre-commit intervention alongside advisory supplements to address conflicts before a pending action is executed.

\looseness=-1
Adding Pulse's execution-time support without RL updates to flat memory improves COMP and Acc by 11.3 and 8.0 percentage points, respectively. With the full memory structure, GRPO improves COMP and Acc by 5.3 points each. After GRPO optimization, delivered memory increases on $\pi$-Bench and decreases on MCP-Persona. Together, these comparisons show complementary roles: structured memories organize user preferences and project context, while streaming support makes this evidence available as task decisions evolve.

\subsection{Supplementary Experiments}
\label{sec:exp-supplementary}

\begin{figure}[!htbp]
  \centering
  \includegraphics[width=\columnwidth]{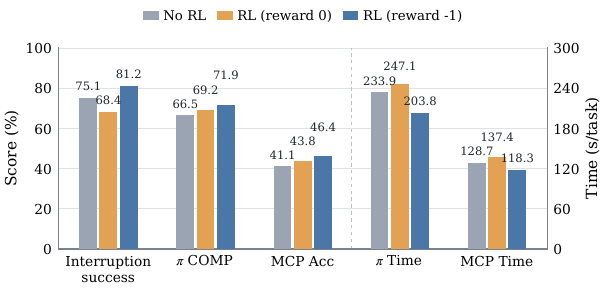}
  \caption{Effects of the incorrect-intervention reward. The $-1$ penalty achieves higher interruption success, $\pi$-Bench COMP, and MCP-Persona Acc than reward $0$. The first three groups use the left axis (\%); the last two use the right axis (s/task).}
  \label{fig:reward-study}
  \Description{Grouped bar chart comparing No RL, RL with reward zero for an incorrect interruption, and RL with reward minus one. Interruption success is 75.1, 68.4, and 81.2 percent, with 494, 557, and 346 total interruptions, respectively; pi-Bench COMP is 66.5, 69.2, and 71.9 percent; MCP-Persona accuracy is 41.1, 43.8, and 46.4 percent. The first three groups use the left axis from 0 to 100 percent. The last two show time on the right axis from 0 to 300 seconds per task: No RL, pi-Bench 233.9 and MCP-Persona 128.7; reward 0, 247.1 and 137.4; reward minus one, 203.8 and 118.3.}
\end{figure}

\begin{figure}[!htbp]
  \centering
  \includegraphics[width=\columnwidth]{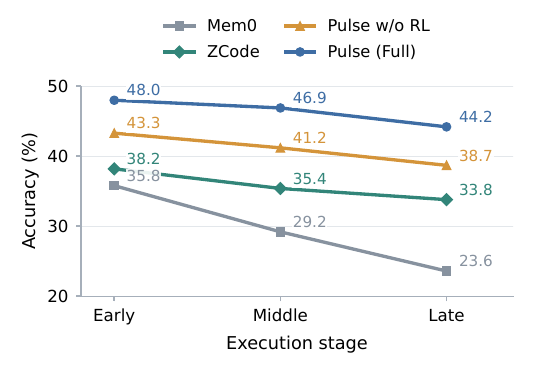}
  \caption{MCP-Persona accuracy across early, middle, and late stages divided by execution step. Pulse (Full) leads at every stage and retains 44.2\% accuracy in the late stage.}
  \label{fig:long-horizon-accuracy}
  \Description{Line chart of MCP-Persona accuracy in early, middle, and late stages divided by execution step. Mem0: 35.8, 29.2, 23.6 percent. ZCode: 38.2, 35.4, 33.8 percent. Pulse without RL: 43.3, 41.2, 38.7 percent. Full Pulse: 48.0, 46.9, 44.2 percent. The vertical axis spans 20 to 50 percent.}
\end{figure}

\begin{figure}[!htbp]
  \centering
  \includegraphics[width=\columnwidth]{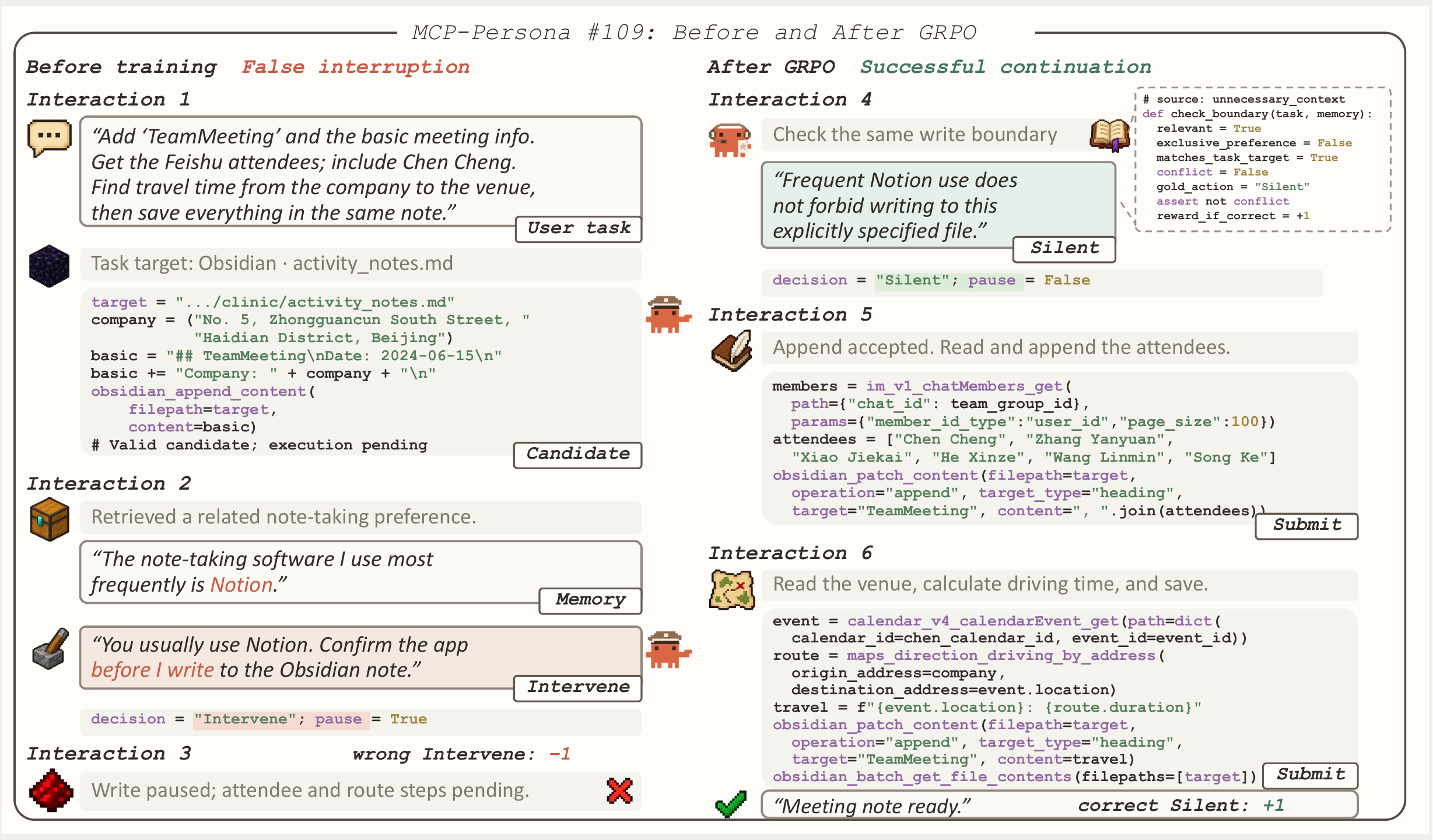}
  \caption{Policy choices in MCP-Persona 109. A habitual Notion preference triggers an unnecessary pause on the left. The right branch selects \textsc{Silent} and continues preparing the requested Obsidian note.}
  \label{fig:mcp109-false-interruption}
  \Description{Six interactions compare two decisions at the same proposed Obsidian append. The left branch treats a frequent-use preference for Notion as an exclusive constraint and pauses a valid write. The right branch recognizes that the memory does not conflict with the task, selects Silent, and continues attendee retrieval, calendar lookup, route calculation, and note updates.}
\end{figure}

\subsubsection{Reward Effects on Intervention Reliability}
\label{sec:exp-reward}

We compare the initial policy with two GRPO~\cite{shao2024deepseekmath} policies that assign an incorrect \textsc{Intervene} a reward of $0$ or $-1$. Correct actions receive $+1$ and other errors receive $0$; the $-1$ setting is Pulse (Full). The RL variants share training data and optimization settings, and all three policies use the same evaluation configuration. Interruption success is the percentage of selected \textsc{Intervene} actions that correctly resolve a memory-dependent conflict. Total interruptions are 494 for the initial policy, 557 for reward $0$, and 346 for reward $-1$.

\looseness=-1
Figure~\ref{fig:reward-study} shows that reward $0$ improves COMP and Acc over the initial policy, but interruption success falls from 75.1\% to 68.4\%. Without a negative reward for incorrect interruptions, the agent lacks an explicit incentive to suppress unnecessary blocking and may learn an overly aggressive intervention strategy. Adding the $-1$ penalty raises interruption success to \textbf{81.2\%}, a gain of 12.8 percentage points over reward $0$, while COMP increases from 69.2\% to 71.9\% and Acc from 43.8\% to 46.4\%. These results support penalizing incorrect interruptions to encourage more selective intervention while improving end-to-end task performance.

\subsubsection{Accuracy over Long-Horizon Execution}
\label{sec:exp-long-horizon}

Figure~\ref{fig:long-horizon-accuracy} compares MCP-Persona accuracy across early, middle, and late stages by execution step: one MCP call or one subtask execution.

\looseness=-1
Pulse (Full) leads throughout, retaining \textbf{44.2\% late-stage accuracy} with a 3.8-point decline, compared with a 12.2-point decline for Mem0. Its late-stage gains over Mem0, ZCode, and Pulse w/o RL are 20.6, 10.4, and 5.5 percentage points, respectively.

\looseness=-1
These trends support more stable personalization as context accumulates and subgoals change. Streaming support supplies evidence for the current decision, while the trained policy helps suppress interruptions based on irrelevant memories. The slower degradation is consistent with improved robustness to historical interference; the advantage over Pulse w/o RL further supports the contribution of policy optimization.

\subsubsection{Case Study}
\label{sec:exp-case}

\looseness=-1
Figure~\ref{fig:mcp109-false-interruption} examines an unnecessary interruption in MCP-Persona task 109. The task requires preparing meeting materials in a specified Obsidian note, including the attendees and travel time. A retrieved memory states that the user most frequently uses Notion. This memory describes a general usage habit, while the current task explicitly specifies Obsidian as the destination.

\looseness=-1
In the left branch, treating the habit as an exclusive constraint leads to \textsc{Intervene}: the valid append is paused for application confirmation, leaving the attendee and route steps pending. At the same decision point, the right branch selects \textsc{Silent}, allowing the task agent to append the meeting section, retrieve attendees and the venue, calculate driving time, and save the materials in the same note.

\looseness=-1
The case demonstrates how Pulse determines memory use in relation to the current candidate action: intervention is based on conflict evidence applicable to that action. Under Equation~\ref{eq:reward}, the unnecessary \textsc{Intervene} receives $-1$, whereas the appropriate \textsc{Silent} receives $+1$. This reward distinction gives GRPO an explicit signal to discourage false interruptions and the replanning they induce, while preserving valid task progress.

\begin{samepage}
\section{Conclusion}
\label{sec:conclusion}

We formulate sustained personalization in long-horizon tasks, examining how memory influences decisions throughout task workflows. Effective personalization depends on when and how historical evidence enters and shapes execution. Pulse combines user and project memories with a memory agent that supplies relevant information to sustain task personalization. GRPO optimizes its choices among silence, supplementation, and intervention. Experiments on multiple datasets show significantly more accurate personalized memory retrieval in both content and timing, alongside more stable performance and higher efficiency throughout long-horizon execution. These findings offer a direction for streaming personalization in agents performing long-horizon tasks.
\end{samepage}

\clearpage
\bibliographystyle{ACM-Reference-Format}
\bibliography{references}

\clearpage
\appendix
\renewcommand{\thetable}{\thesection\arabic{table}}
\setcounter{table}{0}
\section{Method Details}
\label{sec:appendix-method-details}

\subsection{Memory Population and Organization}
\label{sec:appendix-memory-fill}

Memory population takes historical conversations and existing task traces available before the task. Records retain source identities and timestamps, distinguishing user statements, tool receipts, and project artifacts. User information is organized into independently retrievable facts under Topics and subtopics, with entities, attributes, and preference relations attached to the corresponding subgraphs. Entity identities are shared across Topics, and shared entities or factual relations connect relevant domains. For example, programming-language preferences belong to Coding, whereas code examples and response length belong to Writing; their links support joint retrieval when explaining code.

Project memory is isolated by project identity and divided into workflows $\mathcal{W}$ and current information $\mathcal{S}$. Workflows record goals, applicability conditions, steps, tools, and input/output dependencies. Current information records accepted decisions, file and artifact versions, progress, pending items, and stage summaries. Retrieval first selects the user, project, and information cutoff, then matches the request or execution event against Topics and subgoals to recover valid facts, workflows, and project state.

\subsection{Incremental Writes and Updates}
\label{sec:appendix-memory-update}

Updates proceed through candidate extraction, source validation, merging, and commit. Explicitly stated durable preferences and identity facts become user-memory candidates, retaining their supporting quote and event identity. Successfully executed tool calls generate project-state candidates from receipts matched to the proposed action. Candidates are committed together when a conversation closes. Equivalent facts are merged, while new records retain provenance and effective time. An explicit revision establishes a supersession link to the old record, so later retrieval selects the currently valid version. Aborted conversations discard uncommitted candidates.

User preferences, project facts, and workflows have distinct update evidence. User statements confirm preferences; execution results confirm progress and artifact changes; reusable experience is organized into steps and dependencies with applicability conditions. A temporary choice retains its task context, and a later project decision updates the corresponding state record. Retaining sources and version relationships allows subsequent retrieval to establish the relevant user, project, and time interval.

\subsection{Execution-Time Policy Training Protocol}
\label{sec:appendix-training-details}

We generate a new training set using the data-generation scripts provided by Persona2Web~\cite{kim2026persona2web} for GRPO training of Pulse's execution-time policy. Persona2Web first generates implicit preferences; GPT-5.6 sol then constructs trajectory examples and behavior labels from these preference points, including cases that deliberately omit or misstate preferences. Each example contains an observable trajectory prefix, its valid memory snapshot, and a behavior label. Prefixes retain the request, initial context, messages, proposed calls, and tool results; labels distinguish silence, \textsc{Supplement}, and \textsc{Intervene} using the evidence available at that point. Data are split by task, keeping related prefixes in the same partition.

Within each group, we fix the controller state $x_t$ and its valid memory-backend snapshot, and sample $n\geq2$ action-and-evidence outputs $\{o_i\}_{i=1}^{n}$ from the group's frozen behavior policy $\pi_{\mathrm{old}}(\cdot\mid x_t)$. Here $\pi_{\boldsymbol{\theta}}(o_i\mid x_t)$ is the conditional likelihood of the complete candidate response under the fixed access/retrieval protocol; behavior decoding extracts its action field. Rewards follow Equation~\ref{eq:reward}. GRPO uses a batch size of 32 for 100 policy update steps; model configuration appears in Appendix~\ref{sec:appendix-experiment-config}.

The normalized relative advantage for each sampled group is
\begin{equation}
a_i=\frac{r_i-\operatorname{mean}(r_{1:n})}
{\operatorname{std}(r_{1:n})+\epsilon_{\mathrm{num}}},\qquad
\epsilon_{\mathrm{num}}>0.
\label{eq:advantage}
\end{equation}
Here $\epsilon_{\mathrm{num}}$ ensures numerical stability. The conditional likelihood ratio is
\begin{equation}
\rho_i(\boldsymbol{\theta})=
\frac{\pi_{\boldsymbol{\theta}}(o_i\mid x_t)}
{\pi_{\mathrm{old}}(o_i\mid x_t)}.
\label{eq:likelihood-ratio}
\end{equation}
The candidate-level surrogate loss is given in Equation~\ref{eq:grpo}.
Here $\epsilon_{\mathrm{clip}}\in(0,1)$ is the clipping threshold and $\beta\geq0$ the KL coefficient. Unlike the group-generating $\pi_{\mathrm{old}}$, $\pi_{\mathrm{ref}}$ remains an independently frozen initial policy throughout RL. Equal rewards yield zero advantage; KL regularization can still contribute an update.

\section{Experimental Details}
\label{sec:appendix-experiments}

\begin{table*}[t]
\centering
\caption{$\pi$-Bench per-category ablations. Percentages appear above completed/total checkpoints; Store and Retrieve are thousands of tokens per task. Time is seconds per task.}
\label{tab:ablation-details-pi}
\fontsize{7.3}{9}\selectfont
\setlength{\tabcolsep}{2pt}
\renewcommand{\arraystretch}{1.18}
\begin{tabular*}{\textwidth}{@{\extracolsep{\fill}}@{}lp{101pt}rrrrrrrrr@{}}
\toprule
 & & \multicolumn{5}{c}{Checkpoint completeness (\%)} & Overall & \multicolumn{3}{c}{Cost per task} \\
\cmidrule(lr){3-7}\cmidrule(lr){8-8}\cmidrule(l){9-11}
Group & Variant & Researcher & Marketer & Pharmacist & \shortstack{Law\\Trainee} & Financier & COMP & \shortstack{Store\\(k)} & \shortstack{Retrieve\\(k)} & \shortstack{Time\\(s)} \\
\midrule
Memory & Flat Memory & \shortstack[r]{50.0\%\\[1pt]{\fontsize{6.2}{7}\selectfont\textcolor[gray]{0.42}{78/156}}} & \shortstack[r]{48.5\%\\[1pt]{\fontsize{6.2}{7}\selectfont\textcolor[gray]{0.42}{32/66}}} & \shortstack[r]{51.8\%\\[1pt]{\fontsize{6.2}{7}\selectfont\textcolor[gray]{0.42}{57/110}}} & \shortstack[r]{47.7\%\\[1pt]{\fontsize{6.2}{7}\selectfont\textcolor[gray]{0.42}{83/174}}} & \shortstack[r]{51.5\%\\[1pt]{\fontsize{6.2}{7}\selectfont\textcolor[gray]{0.42}{87/169}}} & \shortstack[r]{49.9\%\\[1pt]{\fontsize{6.2}{7}\selectfont\textcolor[gray]{0.42}{337/675}}} & 7.4 & 4.3 & 228.4 \\\addlinespace[3pt]
 & Flat + User & \shortstack[r]{52.6\%\\[1pt]{\fontsize{6.2}{7}\selectfont\textcolor[gray]{0.42}{82/156}}} & \shortstack[r]{51.5\%\\[1pt]{\fontsize{6.2}{7}\selectfont\textcolor[gray]{0.42}{34/66}}} & \shortstack[r]{58.2\%\\[1pt]{\fontsize{6.2}{7}\selectfont\textcolor[gray]{0.42}{64/110}}} & \shortstack[r]{50.6\%\\[1pt]{\fontsize{6.2}{7}\selectfont\textcolor[gray]{0.42}{88/174}}} & \shortstack[r]{55.6\%\\[1pt]{\fontsize{6.2}{7}\selectfont\textcolor[gray]{0.42}{94/169}}} & \shortstack[r]{53.6\%\\[1pt]{\fontsize{6.2}{7}\selectfont\textcolor[gray]{0.42}{362/675}}} & 6.7 & 4.2 & 248.5 \\\addlinespace[3pt]
 & Flat + Project & \shortstack[r]{53.2\%\\[1pt]{\fontsize{6.2}{7}\selectfont\textcolor[gray]{0.42}{83/156}}} & \shortstack[r]{51.5\%\\[1pt]{\fontsize{6.2}{7}\selectfont\textcolor[gray]{0.42}{34/66}}} & \shortstack[r]{60.9\%\\[1pt]{\fontsize{6.2}{7}\selectfont\textcolor[gray]{0.42}{67/110}}} & \shortstack[r]{52.3\%\\[1pt]{\fontsize{6.2}{7}\selectfont\textcolor[gray]{0.42}{91/174}}} & \shortstack[r]{54.4\%\\[1pt]{\fontsize{6.2}{7}\selectfont\textcolor[gray]{0.42}{92/169}}} & \shortstack[r]{54.4\%\\[1pt]{\fontsize{6.2}{7}\selectfont\textcolor[gray]{0.42}{367/675}}} & 6.5 & 4.1 & 245.7 \\\addlinespace[3pt]
 & Flat Memory + Pulse\newline w/o RL & \shortstack[r]{57.1\%\\[1pt]{\fontsize{6.2}{7}\selectfont\textcolor[gray]{0.42}{89/156}}} & \shortstack[r]{53.0\%\\[1pt]{\fontsize{6.2}{7}\selectfont\textcolor[gray]{0.42}{35/66}}} & \shortstack[r]{66.4\%\\[1pt]{\fontsize{6.2}{7}\selectfont\textcolor[gray]{0.42}{73/110}}} & \shortstack[r]{63.8\%\\[1pt]{\fontsize{6.2}{7}\selectfont\textcolor[gray]{0.42}{111/174}}} & \shortstack[r]{62.1\%\\[1pt]{\fontsize{6.2}{7}\selectfont\textcolor[gray]{0.42}{105/169}}} & \shortstack[r]{61.2\%\\[1pt]{\fontsize{6.2}{7}\selectfont\textcolor[gray]{0.42}{413/675}}} & 7.4 & 4.6 & 202.7 \\\addlinespace[3pt]
\midrule
Retrieval & Start-only & \shortstack[r]{54.5\%\\[1pt]{\fontsize{6.2}{7}\selectfont\textcolor[gray]{0.42}{85/156}}} & \shortstack[r]{53.0\%\\[1pt]{\fontsize{6.2}{7}\selectfont\textcolor[gray]{0.42}{35/66}}} & \shortstack[r]{62.7\%\\[1pt]{\fontsize{6.2}{7}\selectfont\textcolor[gray]{0.42}{69/110}}} & \shortstack[r]{54.0\%\\[1pt]{\fontsize{6.2}{7}\selectfont\textcolor[gray]{0.42}{94/174}}} & \shortstack[r]{58.0\%\\[1pt]{\fontsize{6.2}{7}\selectfont\textcolor[gray]{0.42}{98/169}}} & \shortstack[r]{56.4\%\\[1pt]{\fontsize{6.2}{7}\selectfont\textcolor[gray]{0.42}{381/675}}} & 5.3 & 3.5 & 215.7 \\\addlinespace[3pt]
 & Streaming\newline supplement-only & \shortstack[r]{58.3\%\\[1pt]{\fontsize{6.2}{7}\selectfont\textcolor[gray]{0.42}{91/156}}} & \shortstack[r]{53.0\%\\[1pt]{\fontsize{6.2}{7}\selectfont\textcolor[gray]{0.42}{35/66}}} & \shortstack[r]{66.4\%\\[1pt]{\fontsize{6.2}{7}\selectfont\textcolor[gray]{0.42}{73/110}}} & \shortstack[r]{62.6\%\\[1pt]{\fontsize{6.2}{7}\selectfont\textcolor[gray]{0.42}{109/174}}} & \shortstack[r]{64.5\%\\[1pt]{\fontsize{6.2}{7}\selectfont\textcolor[gray]{0.42}{109/169}}} & \shortstack[r]{61.8\%\\[1pt]{\fontsize{6.2}{7}\selectfont\textcolor[gray]{0.42}{417/675}}} & 5.3 & 4.2 & 229.6 \\\addlinespace[3pt]
\midrule
Training & w/o GRPO & \shortstack[r]{61.5\%\\[1pt]{\fontsize{6.2}{7}\selectfont\textcolor[gray]{0.42}{96/156}}} & \shortstack[r]{56.1\%\\[1pt]{\fontsize{6.2}{7}\selectfont\textcolor[gray]{0.42}{37/66}}} & \shortstack[r]{70.0\%\\[1pt]{\fontsize{6.2}{7}\selectfont\textcolor[gray]{0.42}{77/110}}} & \shortstack[r]{70.1\%\\[1pt]{\fontsize{6.2}{7}\selectfont\textcolor[gray]{0.42}{122/174}}} & \shortstack[r]{69.2\%\\[1pt]{\fontsize{6.2}{7}\selectfont\textcolor[gray]{0.42}{117/169}}} & \shortstack[r]{66.5\%\\[1pt]{\fontsize{6.2}{7}\selectfont\textcolor[gray]{0.42}{449/675}}} & 5.3 & 3.6 & 233.9 \\\addlinespace[3pt]
\midrule
 & \textbf{Full Pulse} & \shortstack[r]{\textbf{68.6\%}\\[1pt]{\fontsize{6.2}{7}\selectfont\textcolor[gray]{0.42}{107/156}}} & \shortstack[r]{\textbf{59.1\%}\\[1pt]{\fontsize{6.2}{7}\selectfont\textcolor[gray]{0.42}{39/66}}} & \shortstack[r]{\textbf{80.9\%}\\[1pt]{\fontsize{6.2}{7}\selectfont\textcolor[gray]{0.42}{89/110}}} & \shortstack[r]{\textbf{74.1\%}\\[1pt]{\fontsize{6.2}{7}\selectfont\textcolor[gray]{0.42}{129/174}}} & \shortstack[r]{\textbf{71.6\%}\\[1pt]{\fontsize{6.2}{7}\selectfont\textcolor[gray]{0.42}{121/169}}} & \shortstack[r]{\textbf{71.9\%}\\[1pt]{\fontsize{6.2}{7}\selectfont\textcolor[gray]{0.42}{485/675}}} & 5.3 & 3.9 & 203.8 \\\addlinespace[3pt]
\bottomrule
\end{tabular*}
\end{table*}

\begin{table*}[t]
\centering
\caption{MCP-Persona per-category ablations. Percentages appear above completed/total checkpoints; Store and Retrieve are thousands of tokens per task. Time is seconds per task.}
\label{tab:ablation-details-mcp}
\fontsize{7.3}{9}\selectfont
\setlength{\tabcolsep}{2pt}
\renewcommand{\arraystretch}{1.18}
\begin{tabular*}{\textwidth}{@{\extracolsep{\fill}}@{}lp{101pt}rrrrrrrrrrr@{}}
\toprule
 & & \multicolumn{7}{c}{Checkpoint accuracy (\%)} & Overall & \multicolumn{3}{c}{Cost per task} \\
\cmidrule(lr){3-9}\cmidrule(lr){10-10}\cmidrule(l){11-13}
Group & Variant & CP & CM & SM & EM & Lark & Rednote & \shortstack{Hodge-\\podge} & Acc & \shortstack{Store\\(k)} & \shortstack{Retrieve\\(k)} & \shortstack{Time\\(s)} \\
\midrule
Memory & Flat Memory & \shortstack[r]{31.5\%\\[1pt]{\fontsize{6.2}{7}\selectfont\textcolor[gray]{0.42}{60.5/192}}} & \shortstack[r]{15.3\%\\[1pt]{\fontsize{6.2}{7}\selectfont\textcolor[gray]{0.42}{11.5/75}}} & \shortstack[r]{30.2\%\\[1pt]{\fontsize{6.2}{7}\selectfont\textcolor[gray]{0.42}{26/86}}} & \shortstack[r]{39.2\%\\[1pt]{\fontsize{6.2}{7}\selectfont\textcolor[gray]{0.42}{20/51}}} & \shortstack[r]{27.5\%\\[1pt]{\fontsize{6.2}{7}\selectfont\textcolor[gray]{0.42}{60.5/220}}} & \shortstack[r]{25.0\%\\[1pt]{\fontsize{6.2}{7}\selectfont\textcolor[gray]{0.42}{23/92}}} & \shortstack[r]{6.3\%\\[1pt]{\fontsize{6.2}{7}\selectfont\textcolor[gray]{0.42}{0.5/8}}} & \shortstack[r]{27.9\%\\[1pt]{\fontsize{6.2}{7}\selectfont\textcolor[gray]{0.42}{202/724}}} & 7.3 & 3.1 & 143.2 \\\addlinespace[3pt]
 & Flat + User & \shortstack[r]{33.9\%\\[1pt]{\fontsize{6.2}{7}\selectfont\textcolor[gray]{0.42}{65/192}}} & \shortstack[r]{18.0\%\\[1pt]{\fontsize{6.2}{7}\selectfont\textcolor[gray]{0.42}{13.5/75}}} & \shortstack[r]{33.7\%\\[1pt]{\fontsize{6.2}{7}\selectfont\textcolor[gray]{0.42}{29/86}}} & \shortstack[r]{43.1\%\\[1pt]{\fontsize{6.2}{7}\selectfont\textcolor[gray]{0.42}{22/51}}} & \shortstack[r]{29.1\%\\[1pt]{\fontsize{6.2}{7}\selectfont\textcolor[gray]{0.42}{64/220}}} & \shortstack[r]{26.1\%\\[1pt]{\fontsize{6.2}{7}\selectfont\textcolor[gray]{0.42}{24/92}}} & \shortstack[r]{12.5\%\\[1pt]{\fontsize{6.2}{7}\selectfont\textcolor[gray]{0.42}{1/8}}} & \shortstack[r]{30.2\%\\[1pt]{\fontsize{6.2}{7}\selectfont\textcolor[gray]{0.42}{218.5/724}}} & 6.8 & 2.9 & 158.6 \\\addlinespace[3pt]
 & Flat + Project & \shortstack[r]{32.8\%\\[1pt]{\fontsize{6.2}{7}\selectfont\textcolor[gray]{0.42}{63/192}}} & \shortstack[r]{19.3\%\\[1pt]{\fontsize{6.2}{7}\selectfont\textcolor[gray]{0.42}{14.5/75}}} & \shortstack[r]{32.0\%\\[1pt]{\fontsize{6.2}{7}\selectfont\textcolor[gray]{0.42}{27.5/86}}} & \shortstack[r]{44.1\%\\[1pt]{\fontsize{6.2}{7}\selectfont\textcolor[gray]{0.42}{22.5/51}}} & \shortstack[r]{30.0\%\\[1pt]{\fontsize{6.2}{7}\selectfont\textcolor[gray]{0.42}{66/220}}} & \shortstack[r]{25.5\%\\[1pt]{\fontsize{6.2}{7}\selectfont\textcolor[gray]{0.42}{23.5/92}}} & \shortstack[r]{18.8\%\\[1pt]{\fontsize{6.2}{7}\selectfont\textcolor[gray]{0.42}{1.5/8}}} & \shortstack[r]{30.2\%\\[1pt]{\fontsize{6.2}{7}\selectfont\textcolor[gray]{0.42}{218.5/724}}} & 6.6 & 2.8 & 154.3 \\\addlinespace[3pt]
 & Flat Memory + Pulse\newline w/o RL & \shortstack[r]{38.5\%\\[1pt]{\fontsize{6.2}{7}\selectfont\textcolor[gray]{0.42}{74/192}}} & \shortstack[r]{22.0\%\\[1pt]{\fontsize{6.2}{7}\selectfont\textcolor[gray]{0.42}{16.5/75}}} & \shortstack[r]{38.4\%\\[1pt]{\fontsize{6.2}{7}\selectfont\textcolor[gray]{0.42}{33/86}}} & \shortstack[r]{50.0\%\\[1pt]{\fontsize{6.2}{7}\selectfont\textcolor[gray]{0.42}{25.5/51}}} & \shortstack[r]{36.1\%\\[1pt]{\fontsize{6.2}{7}\selectfont\textcolor[gray]{0.42}{79.5/220}}} & \shortstack[r]{32.1\%\\[1pt]{\fontsize{6.2}{7}\selectfont\textcolor[gray]{0.42}{29.5/92}}} & \shortstack[r]{25.0\%\\[1pt]{\fontsize{6.2}{7}\selectfont\textcolor[gray]{0.42}{2/8}}} & \shortstack[r]{35.9\%\\[1pt]{\fontsize{6.2}{7}\selectfont\textcolor[gray]{0.42}{260/724}}} & 7.3 & 3.2 & 124.4 \\\addlinespace[3pt]
\midrule
Retrieval & Start-only & \shortstack[r]{37.8\%\\[1pt]{\fontsize{6.2}{7}\selectfont\textcolor[gray]{0.42}{72.5/192}}} & \shortstack[r]{20.7\%\\[1pt]{\fontsize{6.2}{7}\selectfont\textcolor[gray]{0.42}{15.5/75}}} & \shortstack[r]{33.7\%\\[1pt]{\fontsize{6.2}{7}\selectfont\textcolor[gray]{0.42}{29/86}}} & \shortstack[r]{49.0\%\\[1pt]{\fontsize{6.2}{7}\selectfont\textcolor[gray]{0.42}{25/51}}} & \shortstack[r]{32.5\%\\[1pt]{\fontsize{6.2}{7}\selectfont\textcolor[gray]{0.42}{71.5/220}}} & \shortstack[r]{26.6\%\\[1pt]{\fontsize{6.2}{7}\selectfont\textcolor[gray]{0.42}{24.5/92}}} & \shortstack[r]{18.8\%\\[1pt]{\fontsize{6.2}{7}\selectfont\textcolor[gray]{0.42}{1.5/8}}} & \shortstack[r]{33.1\%\\[1pt]{\fontsize{6.2}{7}\selectfont\textcolor[gray]{0.42}{239.5/724}}} & 5.3 & 2.2 & 110.4 \\\addlinespace[3pt]
 & Streaming\newline supplement-only & \shortstack[r]{39.6\%\\[1pt]{\fontsize{6.2}{7}\selectfont\textcolor[gray]{0.42}{76/192}}} & \shortstack[r]{22.0\%\\[1pt]{\fontsize{6.2}{7}\selectfont\textcolor[gray]{0.42}{16.5/75}}} & \shortstack[r]{39.0\%\\[1pt]{\fontsize{6.2}{7}\selectfont\textcolor[gray]{0.42}{33.5/86}}} & \shortstack[r]{52.0\%\\[1pt]{\fontsize{6.2}{7}\selectfont\textcolor[gray]{0.42}{26.5/51}}} & \shortstack[r]{35.9\%\\[1pt]{\fontsize{6.2}{7}\selectfont\textcolor[gray]{0.42}{79/220}}} & \shortstack[r]{31.5\%\\[1pt]{\fontsize{6.2}{7}\selectfont\textcolor[gray]{0.42}{29/92}}} & \shortstack[r]{31.3\%\\[1pt]{\fontsize{6.2}{7}\selectfont\textcolor[gray]{0.42}{2.5/8}}} & \shortstack[r]{36.3\%\\[1pt]{\fontsize{6.2}{7}\selectfont\textcolor[gray]{0.42}{263/724}}} & 5.3 & 2.9 & 137.6 \\\addlinespace[3pt]
\midrule
Training & w/o GRPO & \shortstack[r]{43.8\%\\[1pt]{\fontsize{6.2}{7}\selectfont\textcolor[gray]{0.42}{84/192}}} & \shortstack[r]{24.7\%\\[1pt]{\fontsize{6.2}{7}\selectfont\textcolor[gray]{0.42}{18.5/75}}} & \shortstack[r]{44.2\%\\[1pt]{\fontsize{6.2}{7}\selectfont\textcolor[gray]{0.42}{38/86}}} & \shortstack[r]{53.9\%\\[1pt]{\fontsize{6.2}{7}\selectfont\textcolor[gray]{0.42}{27.5/51}}} & \shortstack[r]{41.1\%\\[1pt]{\fontsize{6.2}{7}\selectfont\textcolor[gray]{0.42}{90.5/220}}} & \shortstack[r]{39.1\%\\[1pt]{\fontsize{6.2}{7}\selectfont\textcolor[gray]{0.42}{36/92}}} & \shortstack[r]{37.5\%\\[1pt]{\fontsize{6.2}{7}\selectfont\textcolor[gray]{0.42}{3/8}}} & \shortstack[r]{41.1\%\\[1pt]{\fontsize{6.2}{7}\selectfont\textcolor[gray]{0.42}{297.5/724}}} & 4.7 & 3.3 & 128.7 \\\addlinespace[3pt]
\midrule
 & \textbf{Full Pulse} & \shortstack[r]{\textbf{48.7\%}\\[1pt]{\fontsize{6.2}{7}\selectfont\textcolor[gray]{0.42}{93.5/192}}} & \shortstack[r]{\textbf{30.7\%}\\[1pt]{\fontsize{6.2}{7}\selectfont\textcolor[gray]{0.42}{23/75}}} & \shortstack[r]{\textbf{49.4\%}\\[1pt]{\fontsize{6.2}{7}\selectfont\textcolor[gray]{0.42}{42.5/86}}} & \shortstack[r]{\textbf{61.8\%}\\[1pt]{\fontsize{6.2}{7}\selectfont\textcolor[gray]{0.42}{31.5/51}}} & \shortstack[r]{\textbf{47.0\%}\\[1pt]{\fontsize{6.2}{7}\selectfont\textcolor[gray]{0.42}{103.5/220}}} & \shortstack[r]{\textbf{41.8\%}\\[1pt]{\fontsize{6.2}{7}\selectfont\textcolor[gray]{0.42}{38.5/92}}} & \shortstack[r]{\textbf{43.8\%}\\[1pt]{\fontsize{6.2}{7}\selectfont\textcolor[gray]{0.42}{3.5/8}}} & \shortstack[r]{\textbf{46.4\%}\\[1pt]{\fontsize{6.2}{7}\selectfont\textcolor[gray]{0.42}{336/724}}} & 4.7 & 2.6 & 118.3 \\\addlinespace[3pt]
\bottomrule
\end{tabular*}
\end{table*}

\begin{table}[t]
\centering
\caption{Persona2Web supplementary results. Web EM and Pref R are percentages; Store and Retrieve are thousands of tokens per task. Time is seconds per task.}
\label{tab:persona2web}
\fontsize{7.3}{9}\selectfont
\setlength{\tabcolsep}{2pt}
\renewcommand{\arraystretch}{1.2}
\begin{tabular*}{\columnwidth}{@{\extracolsep{\fill}}@{}lrrrrr@{}}
\toprule
Method & \multicolumn{2}{c}{Performance (\%)} & \multicolumn{3}{c}{Cost per task} \\
\cmidrule(lr){2-3}\cmidrule(l){4-6}
 & Web EM & Pref R & \shortstack{Store\\(k)} & \shortstack{Retrieve\\(k)} & \shortstack{Time\\(s)} \\
\midrule
\multicolumn{6}{@{}l}{\textit{Memory systems}} \\[2pt]
Mem0 & 59.6 & 69.6 & 3.9 & 2.9 & 69.3 \\
LightMem & 34.7 & 54.2 & 1.7 & 1.8 & 62.1 \\
A-MEM & 40.4 & 54.2 & 6.3 & 3.7 & 72.9 \\
\midrule
\multicolumn{6}{@{}l}{\textit{Agent harnesses}} \\[2pt]
OpenClaw & 63.1 & 73.8 & 4.1 & 3.4 & 77.4 \\
dsh & 64.7 & 75.6 & 4.3 & 3.6 & 62.3 \\
ZCode & 65.3 & 76.2 & 4.5 & 3.8 & 75.9 \\
\midrule
\multicolumn{6}{@{}l}{\textit{Pulse variants}} \\[2pt]
Pulse w/o RL & 68.2 & 80.7 & 3.2 & 2.7 & 46.6 \\
\textbf{Pulse (Full)} & \textbf{73.1} & \textbf{86.4} & \textbf{3.2} & \textbf{2.1} & \textbf{44.9} \\
\bottomrule
\end{tabular*}
\end{table}

\begin{table}[t]
\centering
\caption{Shared Pulse memory backend. Performance is in percent; Retrieve is k tokens per task, and Time is seconds per task.}
\label{tab:shared-backend}
\fontsize{7.3}{9}\selectfont
\setlength{\tabcolsep}{2pt}
\renewcommand{\arraystretch}{1.25}
\begin{tabular*}{\columnwidth}{@{\extracolsep{\fill}}@{}lrrr@{}}
\toprule
Method & Performance & \multicolumn{2}{c}{Cost per task} \\
\cmidrule(l){3-4}
 & (\%) & Retrieve (k) & Time (s) \\
\midrule
\multicolumn{4}{@{}l}{\textit{$\pi$-Bench: COMP}} \\[2pt]
Mem0 & 56.4 & 4.7 & 257.0 \\
dsh & 63.7 & 4.5 & 229.2 \\
Pulse w/o RL & 66.5 & \textbf{3.6} & 233.9 \\
\textbf{Pulse} & \textbf{71.9} & 3.9 & \textbf{203.8} \\
\midrule
\multicolumn{4}{@{}l}{\textit{MCP-Persona: Acc}} \\[2pt]
Mem0 & 34.6 & 3.2 & 145.9 \\
dsh & 38.5 & 3.4 & 136.4 \\
Pulse w/o RL & 41.1 & 3.3 & 128.7 \\
\textbf{Pulse} & \textbf{46.4} & \textbf{2.6} & \textbf{118.3} \\
\bottomrule
\end{tabular*}
\end{table}

\subsection{Evaluation Configuration and Training Parameters}
\label{sec:appendix-experiment-config}

We preserve workflow history for $\pi$-Bench and run MCP-Persona tasks independently~\cite{zhang2026pibench,wang2026mcppersona}. COMP and Acc divide completed checkpoints by the corresponding total; partial MCP-Persona completion contributes 0.5. Category and overall percentages are computed from raw counts and rounded to one decimal place. Mem0 uses its original memory backend~\cite{chhikara2025mem0}.

\begin{center}
\small
\renewcommand{\arraystretch}{1.2}
\begin{tabular*}{\columnwidth}{@{\extracolsep{\fill}}@{}>{\raggedright\arraybackslash}p{0.42\columnwidth}>{\raggedright\arraybackslash}p{0.52\columnwidth}@{}}
\toprule
\textbf{Configuration} & \textbf{Value} \\
\midrule
Task agent & GPT-5.4 mini \\
Embedding model & text-embedding-3-small \\
All memory-agent backbones & Qwen3-8B \\
Training data & Newly generated with Persona2Web data-generation scripts \\
Policy optimization & GRPO, 100 update steps \\
Batch size & 32 \\
\bottomrule
\end{tabular*}
\end{center}

Store Token measures the number of tokens output by the memory model to the memory system. Retrieve Token measures the memory tokens supplied by the memory model to the task model, including initial injection, execution-time supplements, and intervention messages; repeated delivery is counted again. Both use the same tokenizer and are averaged per task and reported in k tokens.

Ablations share the task agent, inference budget, initial memory, and writeback rules. Flat Memory serves as the flat-memory baseline; Flat Memory + Pulse w/o RL adds the initial Pulse execution-time policy without RL updates, while w/o GRPO uses the full memory structure and initial policy. Full Pulse, Flat + User, Flat + Project, and Streaming supplement-only share the trained controller and decoding settings. Writeback follows native $\pi$ episodes and task-local MCP boundaries.

\subsection{Memory Evidence in Case Studies}
\label{sec:appendix-memory-hit}

Case-study records connect retrieved evidence, delivered memory, and subsequent actions to distinguish retrieval hits, evidence adoption, and behavioral changes.

\subsection{Per-Category Ablation Results}
\label{sec:appendix-ablation-details}

The following tables report per-category completed checkpoints, percentages, and memory costs for all eight settings.


\subsection{Supplementary Evaluation on Persona2Web}
\label{sec:appendix-persona2web}

Training uses the newly generated data; Table~\ref{tab:persona2web} reports Web EM and Pref R on 450 Persona2Web evaluation tasks, together with Store and Retrieve costs using the memory accounting in Appendix~\ref{sec:appendix-experiment-config}. Pref R is the percentage of the 450 evaluation tasks that satisfy all preferences.

Pulse (Full) achieves 73.1\% Web EM and 86.4\% Pref R, exceeding the strongest non-Pulse baseline, ZCode, by 7.8 and 10.2 percentage points. Compared with Pulse w/o RL, the gains are 4.9 and 5.8 points. Retrieve cost decreases from 2.7k to 2.1k tokens, a 22.2\% reduction, while Store remains at 3.2k. These results indicate that the trained policy improves personalized web-task performance while delivering a more compact memory context.

\subsection{Comparison with a Shared Pulse Memory Backend}
\label{sec:appendix-shared-backend}

Mem0, dsh, Pulse w/o RL, and Pulse share our memory storage and retrieval backend, comprising user Topic graphs and project workflow/state memory. Each run starts from an independent copy of the same memory snapshot, with the embedding model, retrieval interface and budget, writeback rules, task model, memory-model backbone, task set, decoding configuration, and execution budget held fixed. Each method retains its policy for when to access memory and how to supply it during execution.

Table~\ref{tab:shared-backend} reports results for this experiment. On $\pi$-Bench, Pulse achieves 71.9\% COMP with 203.8 seconds per task, compared with 56.4\% and 257.0 seconds for Mem0 and 63.7\% and 229.2 seconds for dsh. On MCP-Persona, Pulse achieves 46.4\% Acc with 118.3 seconds per task, compared with 34.6\% and 145.9 seconds for Mem0 and 38.5\% and 136.4 seconds for dsh.

\clearpage
\end{document}